\documentclass[letterpaper, 10 pt, conference]{ieeeconf} 

\IEEEoverridecommandlockouts 

\usepackage{times}
\usepackage{epsfig}
\usepackage{graphicx}
\usepackage{amsmath}
\usepackage{amssymb}
\usepackage{algorithm}
\usepackage{algorithmic}
\usepackage{subfigure}
\usepackage{cite}

\usepackage[colorlinks,linkcolor=red]{hyperref}

\title{\LARGE \bf
A Registration-aided Domain Adaptation Network for 3D Point Cloud Based Place Recognition
}

\author{Zhijian~Qiao$ ^{1\dagger} $, Hanjiang~Hu$ ^{1,2\dagger} $, Weiang~Shi$^{1}$, Siyuan~Chen$^{1}$, Zhe~Liu$^{3}$, Hesheng~Wang$^{1}$
\thanks{*The first two authors$ ^\dagger $ contributed equally.}
\thanks{$^{1}$Z.~Qiao, H.~Hu, W.~Shi and S.~Chen are with the Department of Automation, Shanghai Jiao Tong University, China. H. Wang is with Department of Automation, Insititue of Medical Robotics, Key Laboratory of System Control and Information Processing of Ministry of Education, Key Laboratory of Marine Intelligent Equipment and System of Ministry of Education, Shanghai Jiao Tong University, Shanghai 200240, China. H. Wang is also with Beijing Advanced Innovation Center for Intelligent Robots and Systems, Beijing Institute of Technology, China.}%
\thanks{$^{2}$H.~Hu is also with the Department of Mechanical Engineering, Carnegie Mellon University, USA (hanjianghu@cmu.edu).}
\thanks{$^{3}$Z.~Liu is with the Department of Computer Science and Technology, University of Cambridge, United Kingdom.}
\thanks{Corresponding author: H.~Wang (wanghesheng@sjtu.edu.cn).}
}

\begin{document}

\maketitle
\thispagestyle{empty}
\pagestyle{empty}

\begin{abstract}
	
	In the field of large-scale SLAM for autonomous driving and mobile robotics, 3D point cloud based place recognition has aroused significant research interest due to its robustness to changing environments with drastic daytime and weather variance. However, it is time-consuming and effort-costly to obtain high-quality point cloud data for place recognition model training and ground truth for registration in the real world. To this end, a novel registration-aided 3D domain adaptation network for point cloud based place recognition is proposed. A structure-aware registration network is introduced to help to learn features with geometric information and a 6-DoFs pose between two point clouds with partial overlap can be estimated. The model is trained through a synthetic virtual LiDAR dataset through GTA-V with diverse weather and daytime conditions and domain adaptation is implemented to the real-world domain by aligning the global features. Our results outperform state-of-the-art 3D place recognition baselines or achieve comparable on the real-world Oxford RobotCar dataset with the visualization of registration on the virtual dataset.

\end{abstract}

\section{Introduction}

In the applications of autonomous driving and mobile robotics, place recognition plays an essential role in the perception and localization. Based on a well-built database of point cloud or images, place recognition aims to retrieve the most similar frames given the query LiDAR scan or camera input. However, since it is of great necessity for the outdoor autonomous vehicle to work from a long-term perspective, the changes of illumination and weather cast a huge challenge for the place recognition. But it has been shown that the point cloud based algorithm is more robust to the changing environmental conditions \cite{angelina2018pointnetvlad,liu2019lpd}.

\begin{figure}[htbp]
	
	\centering
	\includegraphics[width=3.2in]{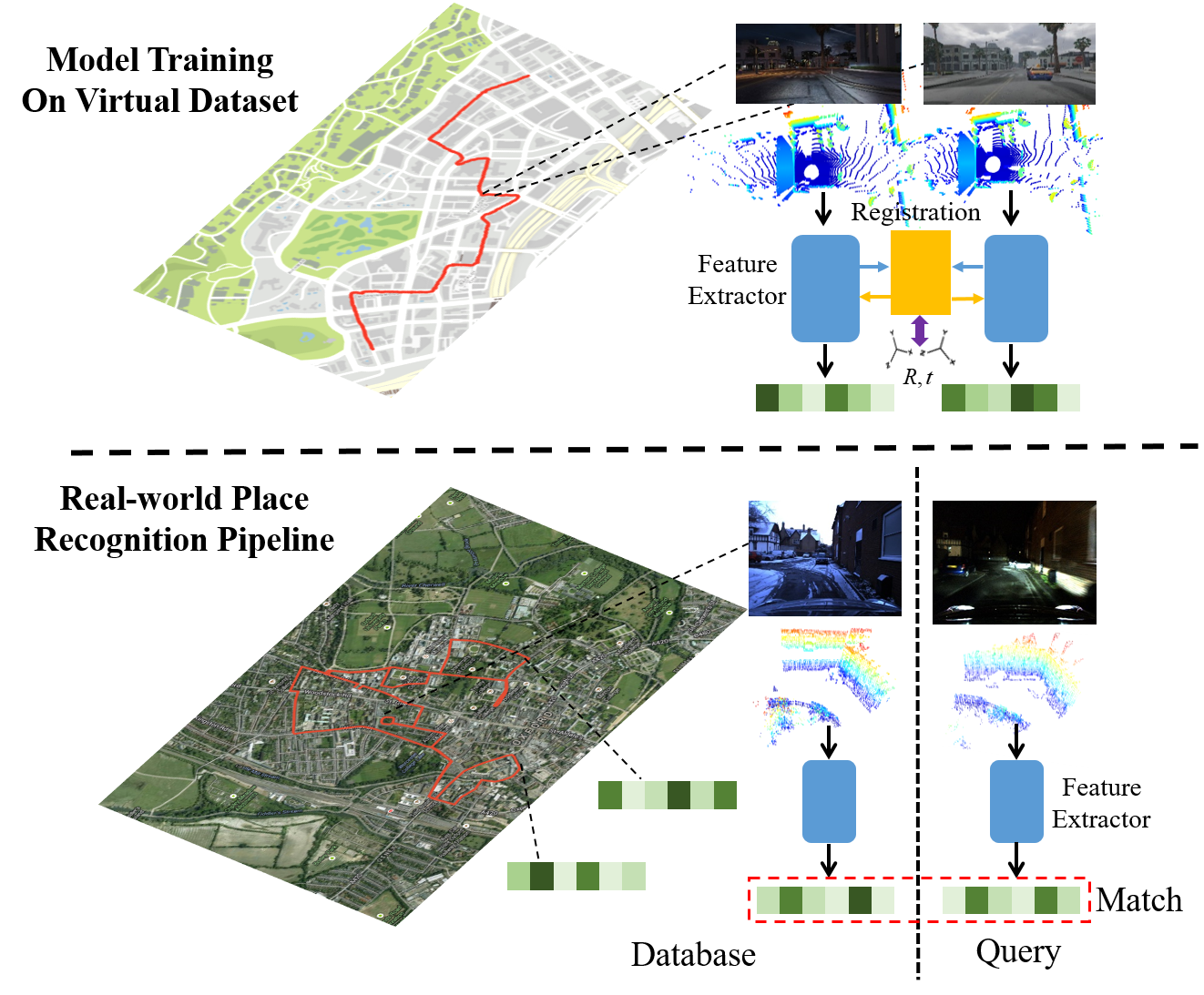}
	\centering
	\caption{The model training and place recognition pipeline is shown above. The model is trained on the virtual dataset with structure-aware feature representation through point cloud registration. For testing, the point cloud with the most similar feature is retrieved given query point cloud feature under changing environments.}
	\label{demo_fig}
\end{figure}

The previous NetVLAD based works on point cloud place recognition \cite{angelina2018pointnetvlad,liu2019lpd} mainly focus on the global feature learned through metric learning. However, for the consideration that global feature can be viewed as the aggregation of local features, we incorporate the point cloud registration to help to learn the local point-wise features for three reasons. First, inspired by \cite{hu2020dasgil}, geometric information can be fused into latent features by a multi-task architecture to learn more discriminative features. Second, some works on partial-to-partial registration \cite{wang2019prnet, 9341249} have shown the ability to extract the structural points of a point cloud, which can be regarded as a method of downsampling. Third, point-wise features can be directly separated for registration to estimate finer pose, which is helpful in both loop closure and re-localization.

In order to further improve performance and generalization ability for point cloud based place recognition, it is indispensable to obtain a large amount of high-quality training data. Nevertheless, it is effort-costly and time-consuming to collect real-world dataset across different types of environments \cite{RobotCarDatasetIJRR}. Consequently, the synthetic virtual dataset has been utilized through GTA-V to involve multi-environment traverses without laborious human efforts, while the accurate registration ground truth can be obtained at the same time.

In this work, we propose a novel registration-aided 3D domain adaptation network with virtual large-scale pointcloud descriptor (vLPD-Net for short) for point cloud based place recognition. A structure-aware registration network is proposed to leverage geometry property and co-contextual information between two input point clouds. Along with them, the adversarial training pipeline is implemented to reduce the gap between synthetic and real-world domain. In the experiment, due to effective domain adaptation from synthetic to real-world datasets for representation learning, our method outperforms state-of-the-art point cloud based place recognition baselines on the Oxford RobotCar dataset. To the best of our knowledge, we are the first to address the outdoor large-scale point cloud based place recognition through domain adaptation with the assistance of point cloud registration. The model training and place recognition pipeline is illustrated in Figure \ref{demo_fig}.

Our contribution can be summarized as follows:
\begin{itemize}
	\item A novel registration-aided domain adaptation model is proposed for the robust large-scale outdoor point cloud based place recognition with feature alignment from synthetic to the real-world domain.
	\item A structure-aware registration network with an efficient outlier removal algorithm is introduced to fuse the geometric information, downsample the raw point cloud and refine the pose estimation.
	\item A multi-environment virtual point cloud dataset is built and the effectiveness of the proposed model is validated through the state-of-the-art performance on the real-world Oxford RobotCar dataset for place recognition and the large-scale virtual dataset for registration with visualization. Our code is available on \url{https://github.com/IRMVLab/vLPD-Net.git}.
\end{itemize}

\section{Related Work}
\label{relatedwork}
\subsection{Point Cloud Based Long-term Place Recognition}
For the point cloud based long-term place recognition task, PointNetVLAD \cite{angelina2018pointnetvlad} is the first end-to-end model to extract global feature of point cloud by combining PointNet and NetVLAD. PCAN \cite{zhang2019pcan} incorporates attention mechanism into PointNetVLAD to learn representation with salient regions. Recent work LPD-Net \cite{liu2019lpd} takes local structural and spatial distribution information into account which achieves state-of-the-art place recognition performance. MinkLoc3D \cite{2021MinkLoc3D} proposes a simple and efficient architecture based on a sparse voxelized point cloud representation and sparse 3D convolutions. However, all the previous works do not consider the geometric structure information and can only obtain coarse retrieve results. 

\subsection{Point Cloud Registration}
Point cloud registration aims to find the rigid transformation from source to target point cloud frames. Traditional ICP-based methods optimize the matching points and the pose transformation, which tend to reach local optimal and are sensitive to the initial pose. Deep neural networks provide more distinguishable features recently. Recent D3feat-NET \cite{bai2020d3feat} detects local features and learns the feature matching simultaneously. DGR \cite{choy2020deep} proposes three modules to 
find pose from a set of putative correspondences accurately and robustly. To fully use the correlation of matching points, Transformer is introduced to find the soft matching points in DCP \cite{wang2019deep}. However, it is difficult and memory-costly to train the attention layers effectively due to a large amount of outdoor point cloud.

\subsection{Unsupervised Domain Adaptation}

Unsupervised domain adaptation aims to address the domain gap and discrepancy between training and the test set. Feature alignment \cite{wang2018low} is implemented to bridge the gap in cross-domain-shared latent feature space. Maximum Mean Discrepancy \cite{borgwardt2006integrating} tries to increase the difference among different classes and decrease the difference inside each class. Domain-invariant features are also used for representation learning to find the shared latent feature space \cite{hu2019retrieval, hu2021domain}. Besides, adversarial training is widely used via discriminator to distinguish the generated features from two domains \cite{ganin2015unsupervised, saito2018maximum}.
Recent work on 3D point cloud data domain adaptation, PointDAN \cite{qin2019pointdan}, proposes a general framework to align both local and global features from source to target domains through adversarial training. However, PointDAN only focuses on object point cloud and does not apply to the whole large-scale outdoor scene with scarcely and unevenly distributed points.


\begin{figure}[]
	\centering
	\includegraphics[width=0.8\linewidth]{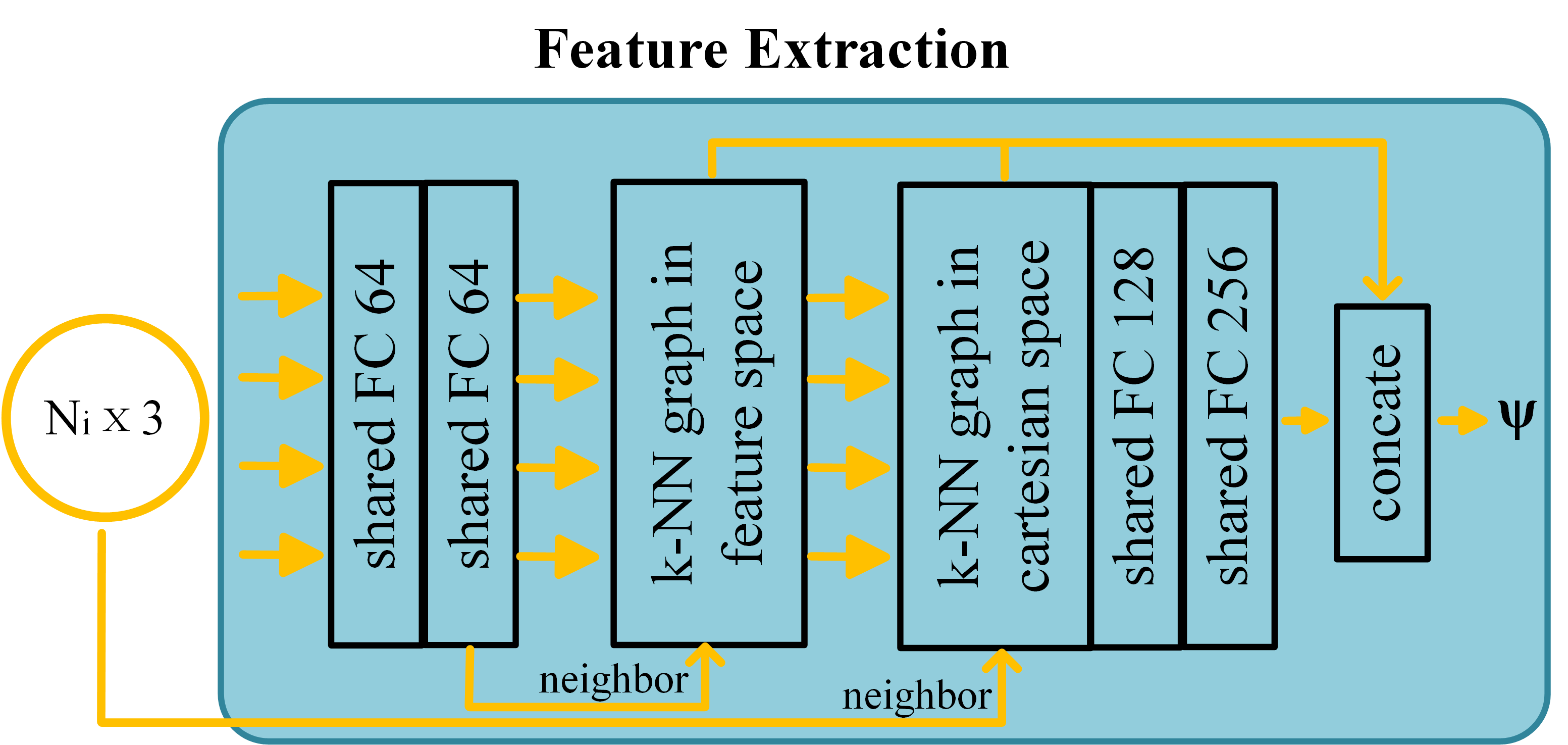}
	\caption{LPD network architecture for feature extraction.}
	\label{lpdnet}
\end{figure}
\begin{figure*}[!t]
	\centering
	\includegraphics[width=2.05\columnwidth]{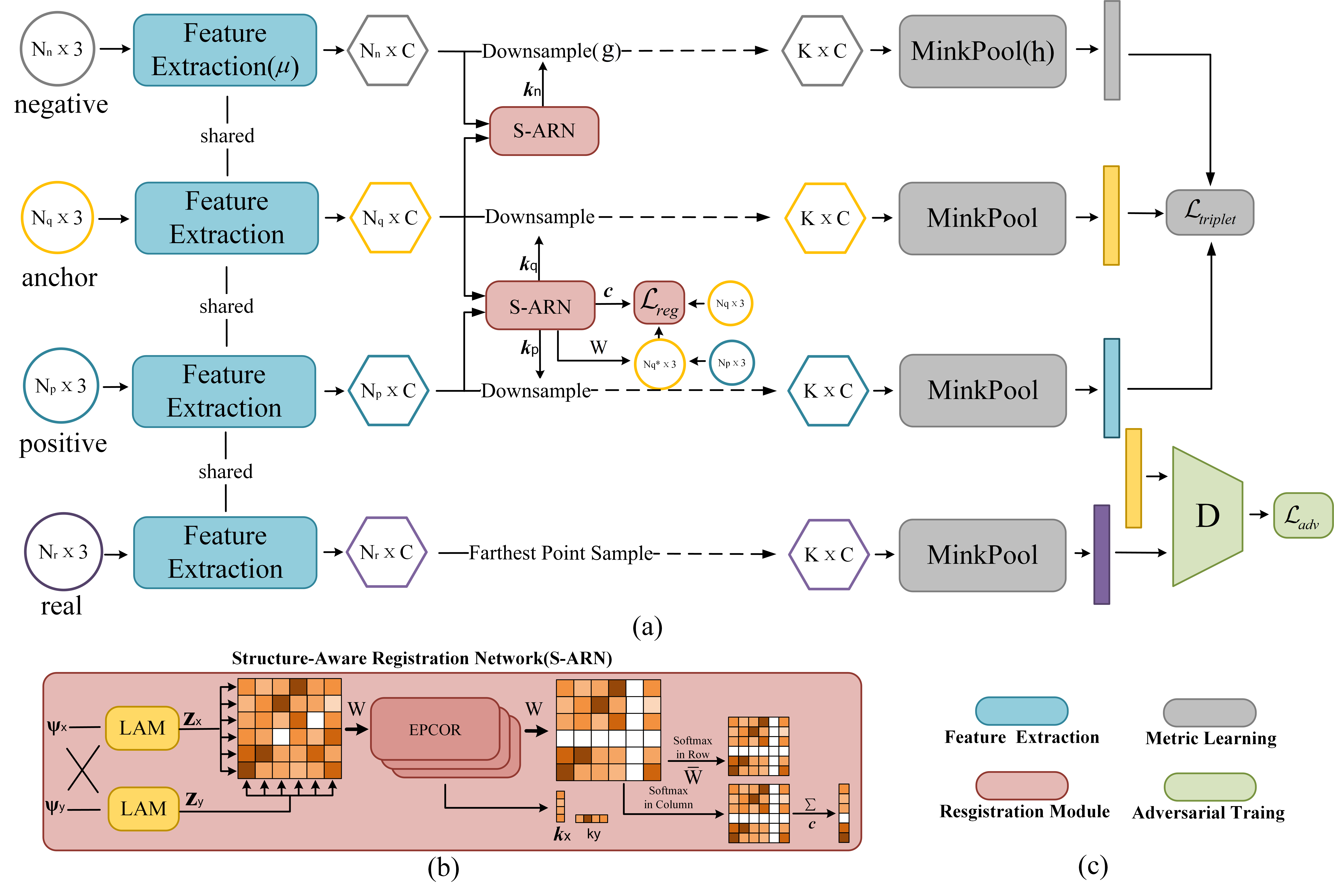}
	\caption{(a) Overview of vLPD-Net. (b) Structure-aware registration network(S-ARN). (c) Colors of different parts about their networks and losses.}
	\label{network}
\end{figure*}

\section{vLPD-Net Model}
\label{architecture}
\subsection{Problem Formulation and System Overview}
\label{sec:overview}
For the problem of point cloud place recognition, the final retrieval is determined through selecting the most similar database candidate given the query point cloud frame. Denote the candidates in the database as 
\begin{align}
	\mathbb{P}_{DB} =& \{ {P_i} \in {{\rm{\mathbb{R}}}^{N{\times}3}} |i = 1,2 \ldots M \} \\ \notag
	{P_i} =& \{ {p_j} \in {{\rm{\mathbb{R}}}^3}|j = 1,2 \ldots {m_i}\} 
\end{align}
where $ M $ is the number of point cloud frames in the database and $ m_i $ is the points in each point cloud frame. 
The problem in this work can be formulated to train the place recognition model on the virtual dataset $\mathbb{P}_{V}$ with the robust feature $ f_V $ for the point cloud frame $P_V$. For testing, the retrieval in the real-world dataset $\mathbb{P}_{R}$ is implemented through the similarity maximization in the database given the query feature $ f_R $. To be concise, the virtual point cloud is denoted as $P$ while the real-world one is $ P_R$. 

The registration-aided domain adaptation architecture is shown in Figure \ref{network}, including feature extraction layer, structure-aware registration network (S-ARN), metric learning module and the adversarial training pipeline.
Given an example of two input point clouds $P$ and $Q$,
\begin{align}
	P &= \left \{ p_{i}\in {\rm{\mathbb{R}}}^3 | i = 1, \cdots, N \right \} \\
	Q &= \left \{ q_{i}\in {\rm{\mathbb{R}}}^3 | i = 1, \cdots, N_{q} \right \}
\end{align}
a feature extraction layer, a structure-aware registration network and a locally aggregated network are sequentially used to encode $P$ and $Q$ into two global descriptor $f_P = h \left ( g \left ( \mu \left ( P \right ), \mu \left ( Q \right ) \right ) \right )$ and $f_Q = h \left ( g \left ( \mu \left ( Q \right ), \mu \left ( P \right ) \right ) \right )$ respectively, where $\mu$ extracts point-wise feature $\psi$, $g$ downsamples $\psi$ to $K$, and $h$ aggregates local features into a global descriptor $f$.

\subsection{Feature Extraction}

Recently, there are many compatible feature extraction layers within the place recognition pipeline. Following \cite{9341249}, we adopt LPD-Net \cite{liu2019lpd} (shown in Figure \ref{lpdnet}) as $\mu$ due to its excellent performance on the large-scale place recognition and registration. It uses graph-based feature aggregation in the Cartesian space as the last feature aggregation operation. And direct connection from later layers to the earlier layers is also adopted to reuse the multi-scale features. Through this layer, $\psi _{P}$ and $\psi _{Q}$ can be obtained from $\mu \left ( P \right )$ and $\mu \left ( Q \right )$ respectively.


\subsection{Structure-Aware Registration Network}
\label{sec:registration}
After neighboring information is well aggregated in the high-dimension space, the point-wise feature goes through a structure-aware registration network to benefit from the spatial and geometric information. The co-contextual information of 3D point cloud pairs is explored for the latent embedding. Inspired from the widely-used Transformer module in natural language processing, the contextual information is encoded through self- $\&$ conditional- attention mechanism for the input embedding in \cite{wang2019deep,9341249}. However, because of the laborious calculation for element-wise correlation distribution, the original Transformer module cost $O(N^2)$ time and memory given the quantity of $N$. 

Based on the observation that the attention distribution from the points of one point cloud to the other tends to be uniform in Transformer under the outdoor large-scale scenes with a large number of points, we propose a light-weighted attention mechanism (LAM), where each point of one point cloud shares the attention weight to the other so that the complexity is $O(N)$ for $N$ points.

Given $z_{P}$ and $z_{Q}$ output from LAM, we define a distance in the feature space to generate the matching matrix $W \in {{\rm{\mathbb{R}}}^{N{\times}N_q}}$. In this work, $\ell^{2}$ distance is used to define each element of $W$ as
\begin{align}
	W\left ( i, j \right ) = \left\| z_{P}^{i} - z_{Q}^{j} \right\|_2, i = 1\dots N, j = 1 \dots N_{q}
\end{align}

After that, a simple but efficient algorithm, named Efficient Point Cloud Outlier Removal (EPCOR), is proposed to remove the non-matching point cloud outliers to obtain sparse matching matrix $\tilde{W}$ and confidence vector $c$ for pairwise registration. The complete algorithm is described in Algorithm \ref{EPCOR}.
\begin{algorithm}[h]
	\caption{Efficient Point Cloud Outlier Removal}
	\begin{algorithmic}
		\label{EPCOR}
		\REQUIRE
		$W \in {\mathbb{R}^{N{\times}N_{q}}}, K\in{\mathbb{R}^{1}}$
		\ENSURE
		$\tilde{W}\in{\mathbb{R}^{N{\times{N_{q}}}}},c\in{\mathbb{R}^{N{\times{1}}}},k,k_q\in{\mathbb{R}^{K{\times{1}}}}$\\
		\STATE $[w_{1}^{T},w_{2}^{T},...w_{N}^{T}]^{T}=W,w_i\in{\mathbb{R}^{1{\times}N_{q}}}$
		\STATE $[m_1,m_2,...,m_{N_{q}}]=W,m_i\in{\mathbb{R}^{N{\times}1}}$

		\FOR{$i \quad in \quad range(N)$}
		\STATE $w_{i}^{'}=softmax(w_{i})$
		\STATE $m_{i}^{'}=softmax(m_{i})$
		\ENDFOR
		\STATE $r=\sum_{i=1}^{N}w_{i}^{'},\quad\quad c=\sum_{i=1}^{N_q}m_{i}^{'}$

		\STATE $k=Index(TopK(c)),\quad k_q=Index(TopK(r))$
		\STATE $W(\complement_Nk,:)=0,\quad W(:,\complement_{N_q}{k_q})=0$
		\STATE $\tilde{W}=softmax(W),\quad c=\frac{c}{\left\| {c} \right\|_1}$
		\RETURN $\tilde{W},c,k,k_q$
	\end{algorithmic}
\end{algorithm}

For each point $p_i$ in $P$, its soft matching point $p_i^*$ is generated with the sparsely-weighted sum of points in $Q$. Each matching pair $(p_i,p_i^*)$ is weighted by the sparse confidence vector $c$, and ${R}$ and ${t}$ can be solved by a weighted Procrustes method in \cite{choy2020deep}. The loss of registration network is denoted as:
\begin{equation}
	\mathcal{L}_{reg} = \frac{1}{\sum_{i=1}^{N} c_i} \sum_{i = 1}^{N}{c_i} \left \| p_i^* - \left ( \hat{R}p_i + \hat{t} \right )| \right \| _2
	\label{reg_loss}
\end{equation}


\subsection{Metric Learning}
In each iteration, the anchor point cloud $P_a$ together with positively paired point cloud $P^+$ and negatively paired point cloud $P^-$ is downsampled by $g$. On the one hand, downsampling can reduce the number of points and speed up network inference. On the other hand, it can remove the noise points and remain the main structure of point cloud. Local features are aggregated into a compact global feature by function $h$ in Section \ref{sec:overview}. To combine the advantages of both voxel-based and point-based strategies \cite{2020PV}, we adopt voxel-based MinkLoc3D \cite{2021MinkLoc3D} to MinkPool as $h$ to aggregate local features more efficiently and effectively while point-based $\mu$ remains accurate point locations with a flexible receptive field via KNN. To simplify the training process, the metric learning loss $\mathcal{L}_{triplet}$ is also consistent with \cite{2021MinkLoc3D}.

\begin{equation}
	\delta^+= \|f(P_a)-f(P^+)\|_2
\end{equation}
\begin{equation}
	\forall P_j^{-}\in {\{P^-\}}, \delta_j^-=\|f(P_a)-f(P_j^-)\|_2 
\end{equation}
\begin{align}
	\label{triplet_loss}
	\mathcal{L}_{triplet}=&\mathop{max}\limits_j([\alpha+\delta^+-\delta_j^-]_+)
\end{align}
where $\alpha$ is a margin hyperparameter.
\subsection{Adversarial Training Pipeline}


\begin{figure*}[htbp]
	\centering
	\subfigure{
		\begin{minipage}[t]{0.13\linewidth}
			\centering
			\includegraphics[width=\linewidth]{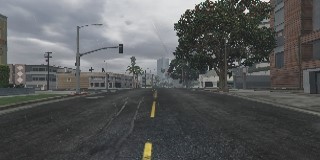}\\
			\vspace{0.1cm}
			\includegraphics[width=\linewidth]{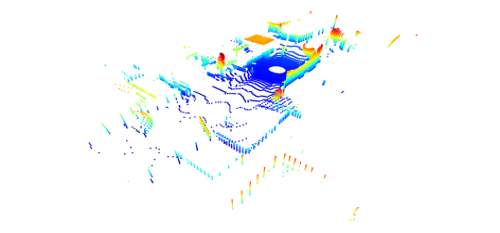}\\
			\vspace{0.1cm}
		\end{minipage}%
	}%
	\subfigure{
		\begin{minipage}[t]{0.13\linewidth}
			\centering
			\includegraphics[width=\linewidth]{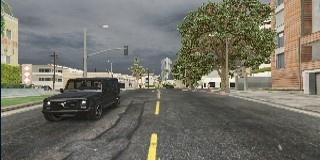}\\
			\vspace{0.1cm}
			\includegraphics[width=\linewidth]{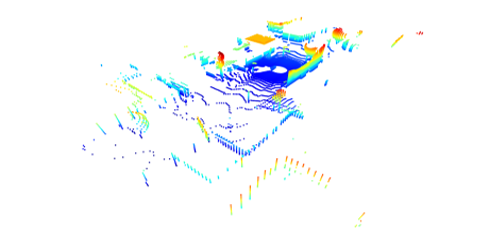}\\
			\vspace{0.1cm}
		\end{minipage}%
	}%
	\subfigure{
		\begin{minipage}[t]{0.13\linewidth}
			\centering
			\includegraphics[width=\linewidth]{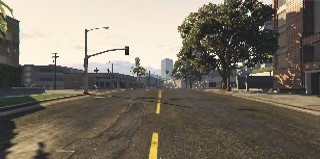}\\
			\vspace{0.1cm}
			\includegraphics[width=\linewidth]{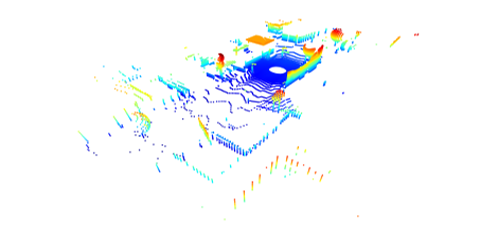}\\
			\vspace{0.1cm}
		\end{minipage}%
	}%
	\subfigure{
		\begin{minipage}[t]{0.13\linewidth}
			\centering
			\includegraphics[width=\linewidth]{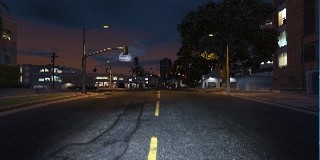}\\
			\vspace{0.1cm}
			\includegraphics[width=\linewidth]{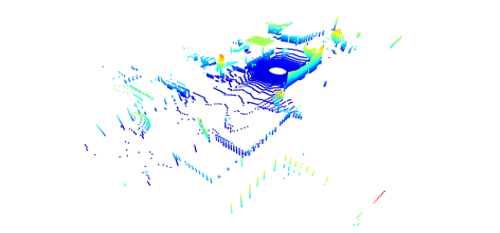}\\
			\vspace{0.1cm}
		\end{minipage}%
	}%
	\subfigure{
		\begin{minipage}[t]{0.13\linewidth}
			\centering
			\includegraphics[width=\linewidth]{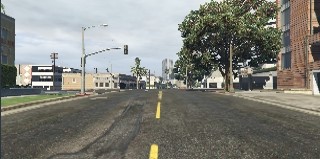}\\
			\vspace{0.1cm}
			\includegraphics[width=\linewidth]{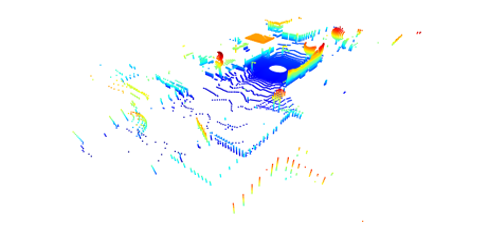}\\
			\vspace{0.1cm}
		\end{minipage}%
	}%
	\subfigure{
		\begin{minipage}[t]{0.13\linewidth}
			\centering
			\includegraphics[width=\linewidth]{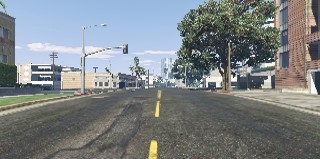}\\
			\vspace{0.1cm}
			\includegraphics[width=\linewidth]{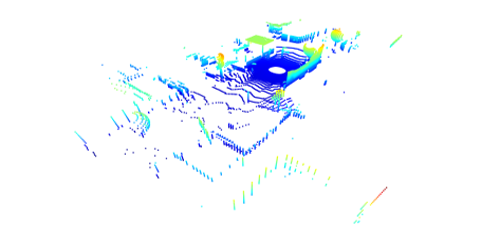}\\
			\vspace{0.1cm}
		\end{minipage}%
	}%
	\subfigure{
		\begin{minipage}[t]{0.13\linewidth}
			\centering
			\includegraphics[width=\linewidth]{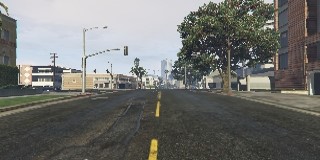}\\
			\vspace{0.1cm}
			\includegraphics[width=\linewidth]{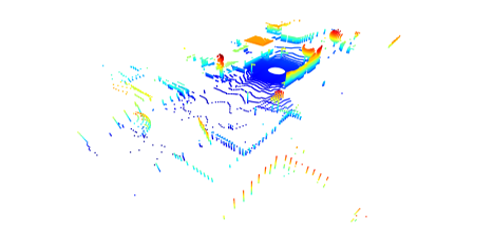}\\
			\vspace{0.1cm}
		\end{minipage}%
	}%
	\centering
	\caption{The RGB and the point cloud samples in the dataset. The environments for each column are Rain, Snow, Dusk, Night, Overcast, Sunny and Cloudy respectively.}
	\label{rgb_pc_samples}
\end{figure*}

Denote virtual point cloud as $P_V \sim p_{V}(P)$ and real-world point cloud as $P_R \sim p_{R}(P)$, and point-wise features from $\mu$ as $\psi_{P_V}$ and $\psi_{P_R}$ accordingly. For the feature from the MinkPool layer, adversarial training is adopted to make the distribution from virtual domain consistent with that from real-world domain. The discriminator is introduced to distinguish the features between virtual and real-world domain while feature extractor layer is optimized in the other step to deceive the discriminator. The adversarial loss consists of two parts, discrimination loss and generation loss:
\begin{align}
	\mathcal{L}_{Dis}&= \mathbb{E}_{ P_{V} \sim p_{V}(P),P_{R} \sim p_{R}(P)}[0.5 \times (D(f(P_{R}))-1)^{2} \notag\\ &+ 0.5 \times D(f(P_{V}))^{2})] \\
	\mathcal{L}_{Gen}&= \mathbb{E}_{ P_{V} \sim p_{V}(P)}[0.5 \times (D(f(P_{V}))-1)^{2}]
	\label{adv_loss}
\end{align}

For the adversarial training, a discriminator $D$ is proposed through a four-layer fully connected network. The optimization of discrimination is instructed by the following step.
\begin{align}
	\mathop {\min }\limits_D {{\cal L}_{Dis}}
\end{align}
The optimization of generation is for the feature extractor, structure-aware registration network and MinkPool module, and denoted as $G$ to include all three networks. The involved losses are registration loss (\ref{reg_loss}), metric learning loss (\ref{triplet_loss}) and generation loss (\ref{adv_loss}), which are shown below. 
\begin{align}
	\label{gen_total_loss}
	\mathop {\min }\limits_G & {\mathbb{E}_{{P_V}\sim{p_V}(P),{P_R}\sim{p_R}(P)}}
	[{\lambda_{triplet}}{{\cal L}_{triplet}}] \notag\\
	+& {\mathbb{E}_{{P_V}\sim{p_V}(P)}}[{\lambda _{reg}}{{\cal L}_{reg}}] + {\cal L}_{Gen}
\end{align}
where $\lambda$ for each term is the weight for the multi-task training in the generation optimization.

\section{Experiment}
\label{experiment}

In this section, we first build a virtual point cloud dataset and specify the experimental setting for preparation. In the experiment, the model is mainly trained on the synthetic virtual dataset, while the real-world dataset is used for the alignment of feature distribution. The comparison experimental results validate the effectiveness of the proposed model. Besides, the virtual large-scale registration results are also compared to other baselines with the promising results.

\begin{figure*}[htbp]
	\centering
	\subfigure{
		\begin{minipage}[t]{0.3\linewidth}
			\centering
			\includegraphics[width=\linewidth,height=0.45\linewidth]{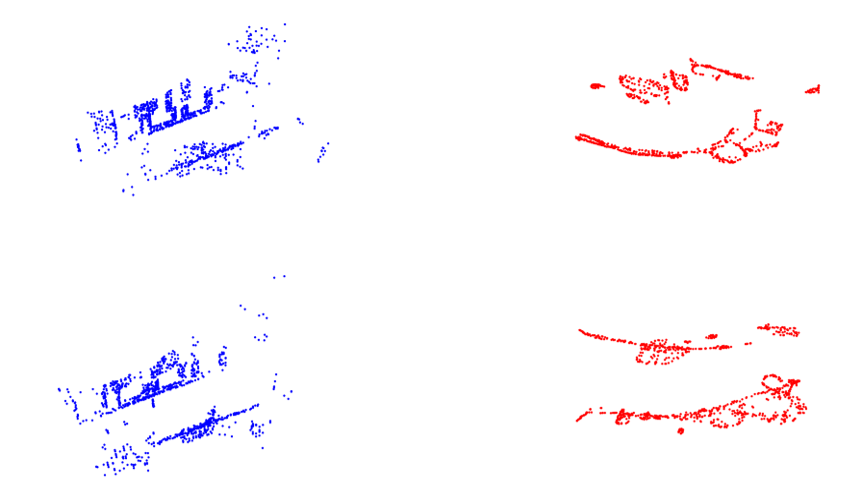}\\
			\vspace{0.1cm}
		\end{minipage}%
	}%
	\subfigure{
		\begin{minipage}[t]{0.1\linewidth}
			\hspace{0.1cm}
		\end{minipage}%
	}%
	\subfigure{
		\begin{minipage}[t]{0.3\linewidth}
			\centering
			\includegraphics[width=\linewidth,height=0.45\linewidth]{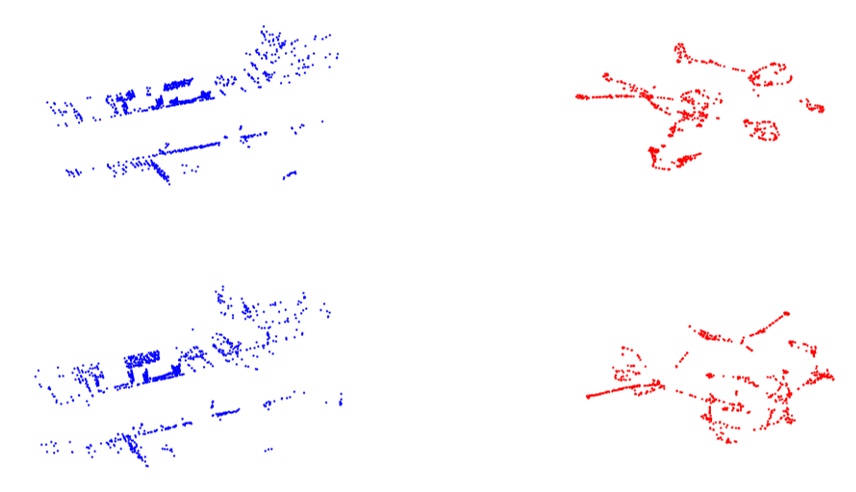}\\
			\vspace{0.1cm}
		\end{minipage}%
	}%
	\centering
	\caption{ The samples of the original point cloud(\textcolor{blue}{blue}) and the t-SNE visualized feature(\textcolor{red}{red}). The left two columns show the results of vLPD-Net with registration network while the right two columns are not with registration network.}
	\label{tSNE_samples}
\end{figure*}

\subsection{Experimental Setup}
To train the model using low-cost and high-quality point cloud traverses under multiple environments, following the previous work \cite{dataset1,dataset2}, we build the virtual synthetic dataset through Grand Theft Auto V (GTA-V) using plugins \cite{url1,url2}. The point cloud frames together with the GPS data are collected via controlling the movement of the car under the game environments with 7 changing weather and daytime along the same route about 2.5km. The collected RGB and point cloud samples are shown in Figure \ref{rgb_pc_samples}. To train the the model effectively for representation learning, the distance between every two frames is set to be 20m, resulting 1,016 virtual point cloud frames under 7 environments. For the loss of point cloud registration, we adopt the ICP algorithm on the basis of GPS as the initial pose to obtain the high-quality rigid transformation as groundtruth. During the process of adversarial training, the real-world dataset Oxford RobotCar \cite{RobotCarDatasetIJRR} is used to align the feature representation. All experiments are conducted on one 2080Ti GPU server.

\subsection{Place Recognition Results}
To evaluate the performance of large-scale place recognition, we use Average Recall@1\% and Average Recall@1 \cite{angelina2018pointnetvlad}. The state-of-the-art baselines we choose are the PointNetVLAD\cite{angelina2018pointnetvlad}, PCAN\cite{zhang2019pcan} and LPD-Net\cite{liu2019lpd}. Moreover, for the fair comparison, we compare the results of LPD-Net and our vLPD-Net using different training datasets, \textit{i.e.} \textit{V} for virtual dataset and \textit{R} for real-world dataset. Note that vLPD-Net \textit{V} is optimized by additional adversarial loss and LPD-Net \textit{R} is pretrained by registration task through training pipline from \cite{9341249}.

\begin{table}[htbp]
	\caption{Experiment Comparison Results on RobotCar Dataset}
	\label{results_pr}
	\begin{center}
		\begin{tabular}{c|c|c}
			\hline
			& Recall@1\% & Recall@1 \\ \hline
			PN-VLAD baseline\cite{angelina2018pointnetvlad} & 81.01 & 62.76 \\
			PN-VLAD refined\cite{angelina2018pointnetvlad} & 80.71 & 63.33 \\
			PCAN\cite{zhang2019pcan} & 86.40 & 70.72 \\
			LPD-Net \textit{R}\cite{liu2019lpd} & 94.92 & 86.28 \\
			vLPD-Net \textit{R} (ours) & \textbf{96.37} & \textbf{89.03} \\ \hline
			LPD-Net \textit{V}\cite{liu2019lpd} & 62.61 & 46.56 \\
			vLPD-Net \textit{V} (ours) & \textbf{73.14} & \textbf{55.43} \\ \hline
		\end{tabular}
	\end{center}
\end{table}

\begin{table}[htbp]
\caption{Registration Comparison Results on the Virtual Dataset}
\label{results_regis}
\centering
\begin{tabular}{c|cccc}
	\hline
	Model & MSE(R) & MAE(R) & MSE(t) & MAE(t) \\ \hline
	ICP & 8.4802 & 1.8171 & 0.0225 & 0.0964 \\
	FGR\cite{2016Fast} & 136.5997 & 7.8012 & 0.0123 & 0.0865 \\
	DCP\cite{wang2019deep} & 2.6111 & 0.9600 & 0.0239 & 0.1024 \\ \hline
	vLPD w/o LAM & 105.5031 & 7.4814 & 0.0288 & 0.1245 \\
	vLPD w/o EPCOR & 9.2840 & 1.9358 & 0.0229 & 0.0964 \\
	vLPD & \textbf{1.9317} & \textbf{0.8451} & \textbf{0.0032} & \textbf{0.0316} \\ \hline
\end{tabular}
\end{table}

From the experimental results of place recognition in Table \ref{results_pr}, LPD-Net \textit{V} presents the worst results because of the overfitting on the perfectly regular point cloud under different environment in the virtual dataset. The performance of vLPD-Net \textit{V} validates the generalization ability of the domain adaptation method compared to LPD-Net \textit{V}. However, the results suck a little bit since the quantity of virtual point cloud frames are much less than real-world frames. vLPD-Net \textit{R} benefits from geometric information from registration and gives the state-of-the-art performance over all the baselines.

\subsection{Registration Results and Visualization}

We conduct another series of experiments to validate the effectiveness of our registration network with compared to several existing approaches, \textit{i.e.} ICP, FGR\cite{2016Fast} and DCP\cite{wang2019deep}(retrained on our dataset). For rotation, the range of the three Euler angles is randomly set from
[-5, 5], [-5, 5] and [-35, 35] respectively. For translation, the range is depended on the retrieved positive pairs. Mean Squared Error (MSE) and Mean Absolute Error (MAE) of rotation drift (R) and translational drift (T) are selected to evaluate the performance of registration.

\begin{figure}[]
	\centering
	\includegraphics[width=0.8\linewidth]{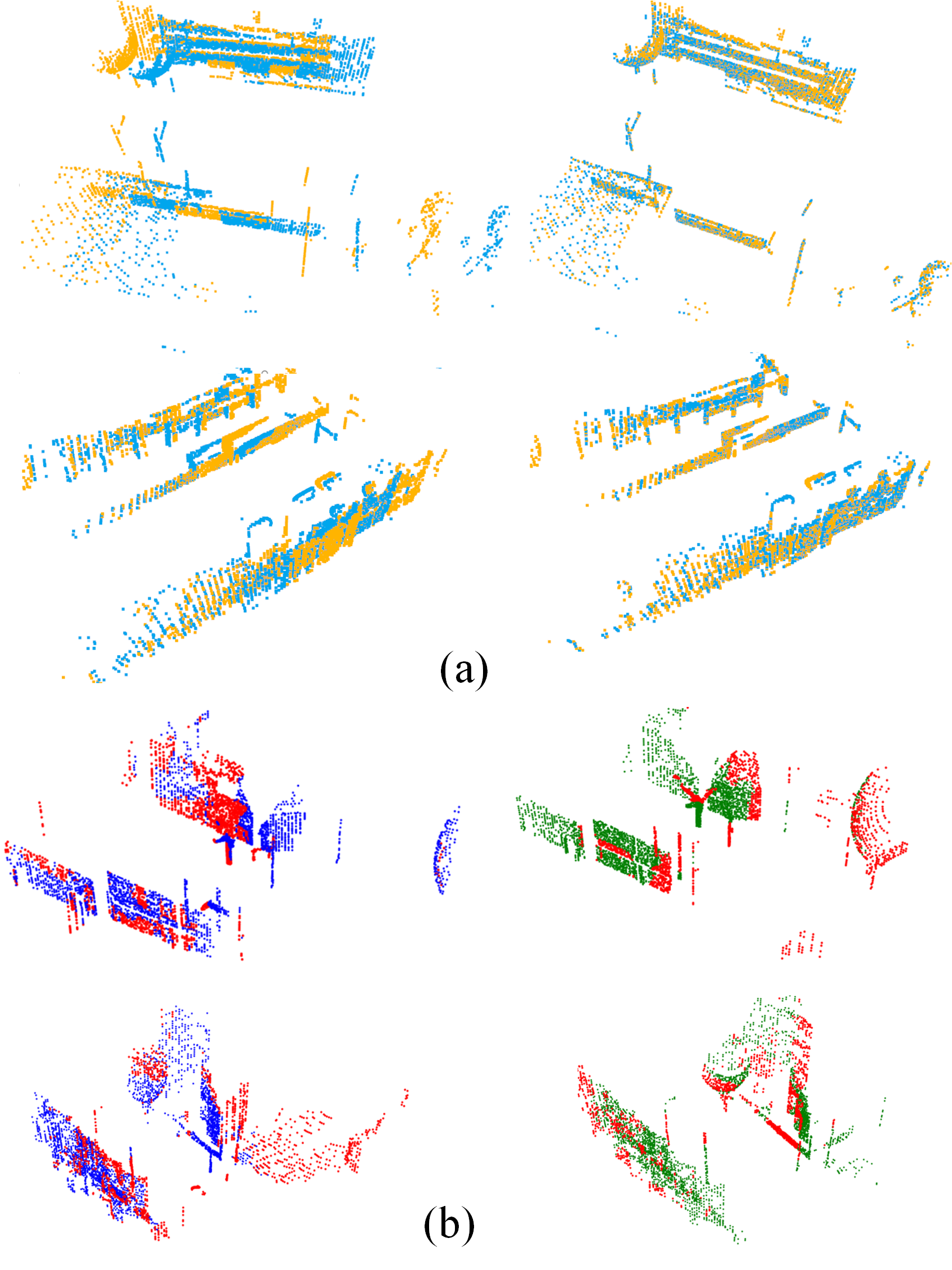}
	\caption{(a) The visualization of point cloud registration on the right column from the initial pose on the left column through the registration network in the vLPD-Net model. (b) The origin retrieved pair is colored with blue and green while outliers are colored with red.}
	\label{reg_samples}
\end{figure}

From the quantitative results in Table \ref{results_regis}, our results outperform all the other. ICP falls into a local optimal without a good initial transformation while FGR fails due to lots of fake matching pairs. Compared to DCP, EPCOR provides an outlier removal method to handle patial-to-patial registration. LAM is proved to make two point clouds knowledgeable about each other and enhance the co-contextual information.
The visualization of the point cloud registration is shown in Figure \ref{reg_samples}, where the vLPD-Net clearly helps to align the point cloud and its retrieved one while outliers are removed by EPCOR.

To explore the influence of structure-aware registration network on feature extraction layer, we visualize the point-wise feature from feature extraction layer by using t-SNE. We compare the dimensionality reduction feature with and without registration network from vLPD-Net in Figure \ref{tSNE_samples}. It can be seen that the correctly-retrieved paired point cloud frames through registration network retain the geometry property in the feature space while the other does not, which validates the structural information can be learned through the guide of registration network.

\section{Conclusion}
\label{conclusion}
In this work, we propose a novel registration-aided domain adaptation network for large-scale point cloud place recognition. A structure-aware registration network is introduced to extract the structural information of the point cloud, which has been visualized in the experiments of point cloud registration. The impressive results of place recognition show that the adversarial training to align the point cloud representation is effective for synthetic-to-real domain adaptation. Finally, it is very promising to learn both global and local features jointly for point cloud retrieval and registration. In the future, we will continue to test its performance on loop closure and re-localization in practical applications.

{\small
	\bibliographystyle{IEEEtran}
	\bibliography{egbib}
}

\end{document}